\pdfoutput=1
\documentclass[11pt]{article}

\usepackage{ACL2023}

\usepackage{times}
\usepackage{latexsym}
\usepackage{multicol}
\usepackage{booktabs}
\usepackage[fleqn]{amsmath}
\usepackage{multirow}
\usepackage[normalem]{ulem}
\usepackage{hyperref}
\usepackage{graphicx}
\usepackage{subfigure}
\usepackage{fixltx2e}
\usepackage{algorithm}
\usepackage{algorithmic}
\usepackage{setspace}
\usepackage{soul}
\usepackage{amsmath}
\usepackage{enumitem}
\usepackage{bm}

\usepackage{tabularx}

\usepackage{tikz}
\usetikzlibrary{tikzmark}
\usepackage[T1]{fontenc}
\usepackage[utf8]{inputenc}

\usepackage{microtype}

\usepackage{inconsolata}

\usepackage{color} % Include the color package
\definecolor{lightgreen}{rgb}{0.6, 0.9, 0.6} % Define a very light green
\definecolor{darkblue}{rgb}{0, 0, 0.5} % This defines a dark blue color

\title{Boosting Biomedical Concept Extraction by Rule-Based Data Augmentation}

\author{Qiwei Shao, Fengran Mo, Jian-Yun Nie\\
DIRO, Université de Montréal, Québec, Canada \\
\texttt{\{qiwei.shao, fengran.mo\}@umontreal.ca, nie@iro.umontreal.ca} \\
}

\begin{document}
\maketitle
\begin{abstract}
Document-level biomedical concept extraction is the task of identifying biomedical concepts mentioned in a given document. Recent advancements have adapted pre-trained language models for this task. However, the scarcity of domain-specific data and the deviation of concepts from their canonical names often hinder these models' effectiveness. To tackle this issue, we employ MetaMapLite, an existing rule-based concept mapping system, to generate additional pseudo-annotated data from PubMed and PMC. The annotated data are used to augment the limited training data. Through extensive experiments, this study demonstrates the utility of a manually crafted concept mapping tool for training a better concept extraction model.
\end{abstract}

\section{Introduction}
Document-level medical concept extraction is crucial for understanding and summarizing medical text. Various models have been proposed, yet there is no consensus on the optimal one. One approach is to model the task as a combination of two simpler subtasks: Named Entity Recognition (NER) and Entity Linking (EL). Many studies treat NER and EL as standalone tasks without integrating them \cite{sung2020biomedical, liu2021fast, shin-etal-2020-biomegatron, 10.1093/bioinformatics/btad103}. This is because either task is challenging due to the lack of data to train for many concepts and the numerous non-canonical forms in which concepts can be written. 

Most medical concept extraction models are constrained by the limited availability of annotated training data, as many medical concepts are rare within training corpora. Existing manually labeled datasets~\cite{li2016biocreative, dougan2014ncbi} are insufficient for training a robust model due to their small size and the absence of many tested concepts within the training corpora. To mitigate this issue, some studies~\cite{DBLP:journals/corr/abs-1901-08746, 10.1145/3458754, DBLP:journals/corr/abs-2010-11784, yasunaga-etal-2022-linkbert, kanakarajan-etal-2021-bioelectra} pre-train language models on unlabeled biomedical texts before fine-tuning them on specific downstream tasks. However, this approach still fails to address the problem of scarce labeling. Other studies incorporate semantic types~\cite{bodenreider2004unified}, entity relations \cite{yuan2021improving, fei2021enriching} and synonyms \cite{sung2020biomedical, liu2020self, miftahutdinov2021medical, yuan2022generative} from biomedical knowledge graphs (KG) to enhance text-to-concept labeling. Despite the enhancements, these methods still rely on external sources for concept-text mapping and fail to consider context during concept extraction, making them less flexible than contextualized approaches.

Early rule-based models have extensively documented the second challenge. These models match text to medical concepts using a predefined dictionary or thesaurus~\cite{aronson2010overview,savova2010mayo,soldaini2016quickumls,demner2017metamap,neumann2019scispacy}. However, such rule-based systems are rigid, only recognizing concepts explicitly covered by the rules. This rigidity poses a problem because many concepts mentioned in documents are not the canonical names stored in dictionaries, making it difficult for rule-based models to correctly classify them. Non-canonical concept names are also challenging for neural models, as they often do not take into account the context in which a concept is mentioned \cite{sung2020biomedical, liu2021fast, angell-etal-2021-clustering, info:doi/10.2196/14830, varma2021cross}. To effectively extract concepts with non-canonical names, the model needs to consider the entire document context. In this paper, we target document-level concept recognition -- a task aiming to recognize the concepts explicitly or implicitly mentioned in a document.

Even though standalone NER and EL models provide valuable insights into simplifying document-level concept extraction, they are limited by assumptions such as the availability of explicit gold-standard concept mentions and reliance solely on mention spans for entity linking. A few pioneering studies attempt to bridge the gap in document-level medical concept extraction by integrating NER and EL into a cohesive pipeline. \citet{aronson2010overview,savova2010mayo,demner2017metamap} utilize syntactic rules to recognize concept mention spans within medical documents and employ a dictionary to map each mention span to a concept ID in the medical ontology. While rule-based models excel at extracting common and standardized concepts, they struggle with rare and non-canonical concept names. \citet{bhowmik-etal-2021-fast} jointly trains NER and EL models initialized from BioBERT \cite{DBLP:journals/corr/abs-1901-08746}, but this approach still depends on gold-standard concept mentions as input for EL, which is not representative of real-world data.
While neural models have shown their effectiveness in recognizing the concepts well covered by the training data, it is known that labeled concepts in the medical domain are limited, making the approach difficult to extend to a wider range of concepts. Our approach leverages an available rule-based concept mapping tool (MetaMapLite) to create more labeled examples to complement the limited training data. We do not assume that the results of such automatic labeling are always correct. Noisy annotations are still useful when the majority of the labels are correct. Our experiments will demonstrate this.

This work tackles two major challenges in document-level medical concept extraction: insufficient training data and non-canonical concept names. The main contributions are as follows:

\begin{itemize}
    \item We propose a pseudo data annotation method using MetaMapLite \cite{demner2017metamap} to address the issue of insufficient training data.
    \item The neural model trained with the augmented data is capable of detecting %We introduce a vector space document-concept matching method to address the challenges posed by 
    non-canonical concept names.
\end{itemize}

\section{Related Work}
%In the domain of Natural Language Processing (NLP), data augmentation plays a crucial role in enhancing model performance by 
Data augmentation aims to generate additional training data for the existing datasets without enough training samples. 
This approach is particularly vital in biomedical NLP tasks, where annotated existing corpora are scarce. 
%Fortunately, the availability of large unannotated biomedical corpora, such as PMC \citep{PMC} and PubMed \citep{PubMed}, provides an opportunity to employ simple heuristics for creating augmented data. 
Several studies~\citep{DBLP:journals/corr/abs-1901-08746, peng-etal-2019-transfer, beltagy-etal-2019-scibert, DBLP:journals/corr/abs-2007-15779, shin-etal-2020-biomegatron, alrowili-shanker-2021-biom, DBLP:journals/corr/abs-2106-03598} have leveraged the large unannotated biomedical corpora, such as PMC \citep{PMC} and PubMed \citep{PubMed} for biomedical data augmentation. 
% For instance, \citet{info:doi/10.2196/14830} have adopted data augmentation techniques inspired by BERT \citep{devlin-etal-2019-bert} to facilitate learning from medical papers and clinical notes. Additionally, approaches proposed by \citet{kanakarajan-etal-2021-bioelectra}, \citet{alrowili-shanker-2021-biom}, and \citet{DBLP:journals/corr/abs-2112-07869} utilize the token corruption technique from ELECTRA \citep{DBLP:journals/corr/abs-2003-10555} to harness insights from unlabeled medical texts.
Beyond the conventional data augmentation techniques employed by general-purpose pre-trained language models, several models have embraced innovative augmentation strategies to enhance their performance. 
%For instance, \citet{liu2021fast} focuses on reconstructing tokens that have been obscured through random span masking and latent layer dropout. 
\citet{varma2021cross} augments entity representation by incorporating entity types and descriptions sourced from Wikipedia. \citet{yasunaga-etal-2022-linkbert} leverages citation links to capture the relationships between PubMed articles. \citet{wang2022improving} boosts sentence representation by integrating documents retrieved from search engines. \citet{tang2023does} employs ChatGPT to create synthetic training examples, while \citet{Liu_Sun_Li_Wang_Zhao_2020} utilizes fuzzy matching to generate pseudo annotations. Similarly, 
\citet{BARTOLINI2023102291} enhances the training dataset by replacing entity mentions with similar terms, and \citet{li2021weakly} trains models to adhere to simple, manually crafted logical rules for Named Entity Recognition (NER).

A notable trend is the incorporation of medical knowledge, such as UMLS \citep{bodenreider2004unified}, into the data augmentation process. This approach is valuable for its rich entity descriptions and structured relationships. Both \citet{yuan2021improving} and \citet{fei2021enriching} use graph neural networks to learn the intrinsic structure of medical knowledge graphs. Moreover, \citet{sung2020biomedical}, \citet{liu2020self}, \citet{miftahutdinov2021medical}, and \citet{yuan2022generative} utilize synonyms for identical concepts as defined in UMLS to familiarize the models with the various nomenclatures of medical entities. 
% \citet{zhang2021knowledge} capitalizes on the context surrounding medical entities in PubMed to enhance entity representation. 
\citet{michalopoulos-etal-2021-umlsbert} integrates semantic types from UMLS into the BERT framework. These works showcase the evolving landscape of data augmentation in NLP where domain-specific knowledge and innovative techniques converge to significantly advance model understanding and performance. Unlike previous approaches, our data augmentation method generates additional training data specifically for document-level concept extraction, demonstrating greater effectiveness compared to knowledge infusion from different tasks.

Our work deals with document-level medical concept extraction -- a task of identifying unique medical concepts across an entire document, irrespective of their specific locations within the text. This task is important for medical information retrieval trying to identify documents on specific medical concepts. Traditionally, this task is formulated as a pipeline involving two tasks: NER and EL. The majority of previous studies focus narrowly on one of the two tasks. Notably, \citet{pmlr-v85-sachan18a}, \citet{DBLP:journals/corr/abs-1901-08746}, \citet{10.1145/3458754}, \citet{yasunaga-etal-2022-linkbert}, and \citet{kanakarajan-etal-2021-bioelectra} have primarily focused on enhancing NER, whereas works such as \citet{DBLP:journals/corr/abs-2010-11784}, \citet{yuan2022biobart}, and \citet{bodenreider2004unified} have primarily focused on EL. A few studies have attempted to fully investigate both steps in the pipeline. \citet{savova2010mayo}, \citet{aronson2010overview}, \cite{soldaini2016quickumls}, \cite{demner2017metamap} use rule-based NER and dictionary-based EL to extract concepts from documents. \citet{bhowmik-etal-2021-fast} uses BERT instead for NER and EL. This NER-then-EL pipeline depends heavily on fine-grained annotations of concept mention spans and struggles with concepts stated in their non-canonical names, as illustrated in Table \ref{Table:non_can_ex}, due to insufficient training data and limited mention context window. Our model addresses the issue by training a more robust model with augmented labeled data, which exposes the model to more situations where concepts are used. % using the whole document as context.

% A few studies have attempted to fully investigate end-to-end EL. \citet{leaman2016taggerone} utilizes a semi-Markov model for NER and compares vector similarity for NEN.  \citet{neumann2019scispacy} develops an entity linking framework that integrates a transition-based NER model with a TF-IDF based NEN model. \citet{safranchik2020weakly} \citet{aronson2010overview} use a lexical-variant retriever and weighted reranker to link text to entities. \citet{demner2017metamap} offer a compact alternative leveraging dictionary-based entity linking. \citet{savova2010mayo} relies on part-of-speech (POS) tagging and syntactic parsing for NER and matches mentions to concepts using a lexicon. \citet{soldaini2016quickumls} first identifies all possible text spans by length before mapping spans to concepts with Jaccard similarity. \citet{vashishth2021improving} adds semantic type filtering to ScispaCy \citep{neumann2019scispacy}, MetaMap \citep{aronson2010overview}, MetaMapLite \citep{demner2017metamap}, cTAKES \citep{savova2010mayo} and QuickUMLS \citep{soldaini2016quickumls} to accurately match mentions to their appropriate entity types. \citet{bhowmik-etal-2021-fast} utilize a BERT-inspired model to pinpoint the most probable entity mention across various spans and subsequently align the mention with its corresponding entity. Inspired by both neural and rule-based methods, our approach gets the best of both worlds by fine-tuning a neural language model on pseudo-annotations provided by a rule-based method.

%\section{Document-level Medical Concept Extraction Task}
\section{Task Definition: Document-level Medical Concept Extraction}
Document-level medical concept extraction task aims to identify unique medical concepts across an entire document, irrespective of their specific locations within the text. 
%This task is crucial as many question answering models can be enhanced by understanding the concepts present in each query and document \cite{nentidis2020results}, without needing the exact mention spans.  
The extracted information can be utilized by various downstream tasks, e.g., enhancing the biomedical matching between each query and document by the presented concepts~\cite{nentidis2020results} rather than the exact mention spans.  
%Traditionally, research in this field has focused narrowly on either Named Entity Recognition (NER) or Entity Linking (EL), overlooking the holistic approach to document-level concept extraction. They have missed the bigger picture of treating the document-level concept extraction task coherently. Although it is possible to model this task as a two-step process combining NER and EL, such a method depends heavily on fine-grained annotations of concept mention spans and struggles with implicitly mentioned concepts where the document references a concept without mentioning its canonical name (Figure \ref{Figure:implicit}). Such shortcomings are typically due to insufficient training data and limited mention context window. 

\begin{table}[h]
\centering
\begin{tabularx}{\linewidth}{X}
\toprule
\textbf{Text:} Abnormal hepatic copper accumulation is recognized as an \underline{inherited disorder} in man, mouse, rat, and dog. \\
\textbf{Concept Mention:} inherited disorder \\
\textbf{Canonical Concept Names:} Hereditary Diseases, Genetic Disorders, syndrome genetic disorder, molecular disease \\
\bottomrule
\end{tabularx}
\caption{Example of the concept ``Hereditary Diseases'' mentioned in its non-canonical form ``inherited disorder''.}
\label{Table:non_can_ex}
\end{table}

Unlike the traditional NER-then-NEN approach~\cite{bhowmik-etal-2021-fast}, we reformulate document-level medical concept extraction as a retrieval problem to gain a holistic view of the document when extracting concepts. The goal is to identify distinct biomedical concepts within a document based on similarity. Formally, given a concept collection $\mathcal{C}$ and an input document $\mathcal{D}$, the model $f$ must retrieve the top-$k$ candidate concepts $c_k$ as follows:
\begin{equation}
\label{eq: task}
    c_k = \mathop{\arg\max}\limits_{k}f(\mathcal{D}, e_i), \ e_i \in \mathcal{C} \ \text{and} \ 0 \leq i \leq k 
\end{equation}
%we train a document embedding model $f(X)$ to find the top $k$ candidate concepts such that:
%\begin{equation}
%    \text{sim}(f(X), \theta e_{i})  \geq \text{sim}(f(X), \theta e_{j}), \forall i, \ 1 < i \leq k \ \text{and} \ \forall j, \ j > k 
%\end{equation}

%where $f(X)$ is a trainable model for document embedding, $\theta e$ is the precomputed concept embedding and sim is the similarity function. We opt for pre-computed concept embeddings instead of trainable ones because there is a large collection of concepts and our computing hardware cannot handle so much back propagation. We use cosine similarity as the sim function:
%\begin{equation*}
%    \text{sim}(f(X), \theta e_{i}) = \frac{f(X) \cdot \theta e_{i}}{\|f(X)\| \|\theta e_{i}\|}
%\end{equation*}

%The trainable document embedding model $f(X)$ follows the same architecture as BioBERT, BiomedBERT, SapBERT and BioLinkBERT, so that we can focus on the effect of data augmentation alone.

\section{Method}
\subsection{Data Augmentation}
%Benchmark datasets of document-level concept extraction lack training instances for the vast majority of concepts in the test data. 
The lack of training examples in document-level concept extraction datasets leads to poor performance of the existing systems, especially for those concepts that need to be recognized but do not occur in training data.
The main reason is the low generalization capacity of the systems trained with limited data.
Thus, the addition of pseudo-annotated training examples for low-resource concepts can mitigate such issues and improve overall performance.
To increase the number of training examples of rare entity labels, we propose an augmentation strategy, which involves candidate document retrieval, pseudo document annotation, and post-annotation filtering.
%This has resulted in poor performance of even the state-of-the-art transformer models. Our exploratory study shows that the performance of transformer is extremely sensitive to the number of training instances. 
%Particularly, the concepts that lack training instances get an F1 close to 0, while those with abundant training instances get a much higher F1. We think the addition of pseudo-annotated training instances for low-resource concepts can mitigate the polarized performance and improve the overall performance. 
%To increase the number of training instances of rare entity labels, we propose an augmentation strategy to solve the data scarcity. Figure \ref{Figure:aug} provides an overview of the data augmentation strategy, which involves candidate document retrieval, pseudo document annotation, and post-annotation filtering.

\subsubsection{Candidate Document Retrieval} 
\label{Sec:retrieval}
We aim to improve the performance of rare concept prediction by generating additional training examples exclusively for concepts below a predetermined threshold of occurrences in the training data\footnote{We also tried to augment training data for all concepts, but the strategy was less successful due to the noise introduced by pseudo-labeling.}.
The first step is identifying the canonical names of these concepts in UMLS \cite{bodenreider2004unified}. If a concept has several canonical names, we use its top canonical name as per UMLS, which is considered the most common name in English. 
Then, the canonical names serve as the search queries in the PMC \cite{PMC} and PubMed \cite{PubMed} to find the candidate documents that contain this concept. 
%We use the default search function on PMC and PubMed, which normalizes the phrase into lower case, and translates maps the query to a MeSH \cite{lipscomb2000medical} concept according to the MeSH translation table. 
% To ensure we have enough training instances after the filtering step, 
The top 50 candidate documents from the search are stored so that we can investigate how the number of augmented documents affects the performance. 
To prevent data leakage, all candidate documents that overlap with the target dataset are excluded. 
Documents are divided into segments with a maximum of 512 tokens where each segment is a training example to comply with the length requirement of transformer models.
%After retrieving the documents, we divide them into segments at the punctuation mark before reaching 512 sub-word tokens. Each segment constitutes a training instance to make sure the length of each instance does not exceed the limit of the transformer models.

\begin{figure*}[!t]
\centering
\includegraphics[width=0.9\linewidth]{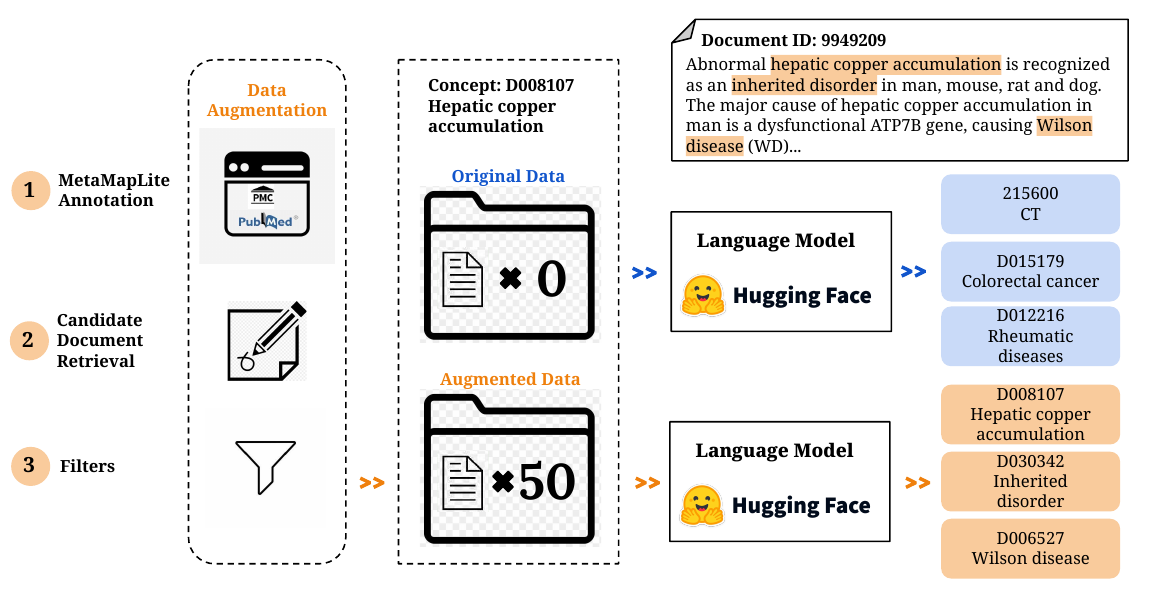}
    \caption{Concept extraction model with data augmentation. The concept \textbf{Hepatic copper accumulation} gets additional training data, and helps the extraction of this concept during inference.}
\label{Figure:aug}
%\vspace{-2ex}
\end{figure*}

\subsubsection{Pseudo Document Annotation}
To create more annotated examples of rare concepts, we use MetaMapLite \cite{demner2017metamap} to annotate these candidate documents, which relies on a rule-based pipeline. 
%MetaMapLite relies on a pipeline of rule-based tokenization and term normalization, POS tagging, dictionary lookup and negation detection to link text to medical concepts. Given an input text, MetaMapLite normalizes the text by removing parentheticals, undoing the inversion, converting the characters to lower-case and stripping the possessives, before looking up the text in a dictionary of UMLS concepts. 
The output format contains the detected concept mention span and a list of potential concepts, ranked by a text-matching score. Our method selects only the highest-ranked concept for each mentioned span. All the concept mention spans in the output of MetaMapLite are kept until the filtering process mentioned in Section \ref{Sec:overlap}.

The concepts in the MetaMapLite output are represented by UMLS concept IDs, which differ from the OMIM and MeSH concept IDs used in our annotated target data, NCBI-Disease, and BC5CDR. This requires a conversion from the UMLS concept IDs to the OMIM and MeSH IDs using the inherent mappings provided by UMLS. Notably, some UMLS IDs can correspond to multiple MeSH and OMIM IDs. In such cases, we select the ID associated with a canonical name that appears in the annotated document.

To capitalize on the close relation between the number of labeled training documents and model performance in our preliminary experiment, our pseudo-annotation process makes sure the low-resource concepts have at least a number of training documents that should render a reasonable performance according to our preliminary experiments. We determine a training occurrence threshold of $k$. For every entity whose training occurrence is below $k$ in the manually annotated training data, our system adds training examples annotated by MetaMapLite to reach the required %until the number of its combined training occurrence in the corpus reaches 
$k$ documents. Then the documents in the corpus are shuffled to ensure that each training batch contains a randomized mix of manually annotated and pseudo-annotated documents.

\subsubsection{Post-annotation Filters}
The output of MetaMapLite often struggles with abbreviations and does not always comply with the annotation protocol of the BC5CDR and NCBI datasets. To address this issue, we apply the following filters to the annotations of MetaMapLite before adding them to the training data.

% \begin{table}
% \centering
% \begin{tabular}{lc}
% \toprule
%      Filter &  \% of Filtered Examples\\
%       \midrule
%       False Abv. & 5.47\\
%       Overlap & 8.5\\
%       Diversity & 34.56\\
%         \bottomrule
% \end{tabular}
% \caption{\label{Table:filtered}
% Percentage of PubMed documents removed by the filters}
% \end{table}

\paragraph{False Abbreviation Filter} 
\cite{demner2017metamap} has noted that the annotations made by MetaMapLite are not on the same level as manual annotations. 
A typical mistake is MetaMapLite incorrectly associate a lower-case word to its upper-case counterpart which is an abbreviation of a medical entity.
% After analyzing the output of MeaMapLite further, we find 5.47\% of the annotations by MetaMapLite incorrectly associate a lower-case word to its upper-case counterpart which is an abbreviation of a medical entity (Table \ref{Table:filtered}). 
Tableau \ref{Table:false_abbv} shows an example of false annotation. MetaMapLite considers the word ``was'' a medical entity because its upper-case form ``WAS'' is an abbreviation of Wiskott-Aldrich Syndrome. To handle mistakes like this, we remove all the abbreviated annotations that do not appear in the document in the upper-case form.

\begin{table}[h]
\centering
\begin{tabularx}{\linewidth}{X}
\toprule
\textbf{Input:} This gene defect was shown to cause a null allele as the result of complete intron retention. \\
\textbf{MetaMapLite Output:} \{Mention: was, Concept ID: C0043194, Concept Name: Wiskott-Aldrich Syndrome\} \\
\bottomrule
\end{tabularx}
\caption{Example of false abbreviation by \textbf{MetaMapLite}.}
\label{Table:false_abbv}
\end{table}

\paragraph{Overlapping Annotation Filter} \label{Sec:overlap}
Another common mistake by MetaMapLite is it often recognizes both the broad and the narrow concepts mentioned in a sentence. For instance, both ``chronic kidney disease'' and ``disease'' are included in the MetaMapLite annotations given the sentence in Tableau \ref{Table:overlap}. 
% This accounts for 8.5\% of the filtered annotations (Table \ref{Table:filtered}).

\begin{table}[h]
\centering
\begin{tabularx}{\linewidth}{X}
\toprule
\textbf{Input:} Signs and symptoms of APRT deficiency caused by stone formation in the kidney that caused obstruction, infection, or chronic kidney disease. \\
\textbf{MetaMapLite Output:} \{Mention: chronic kidney disease, Concept ID: C1561643, Concept Name: Chronic Kidney Disease\};  \{Mention: disease, Concept ID: C0012634, Concept Name: Disease\}\\
% \textbf{Output 2:} \{Mention: disease, Concept ID: C0012634, Concept Name: Disease\}\\
\bottomrule
\end{tabularx}
\caption{Example of overlapping annotations by \textbf{MetaMapLite}.}
\label{Table:overlap}
\end{table}

Contrary to MetaMapLite, our task data would only include the most specific concept of the two, ``chronic kidney disease''. To make the augmented data align with the protocol of the task data, we remove any annotation of a broad concept that has an overlapping span with a more specific concept in the sentence.

\paragraph{Diversity Filter}
Different papers usually use different names when they refer to the same concept. The model has to learn a variety of concept names from various contexts, but many retrieved candidate documents are duplicated. 
% However, 34.56\% of the candidate documents retrieved in Section \ref{Sec:retrieval} are duplicated (Table \ref{Table:filtered}). 
This motivates us to employ a diversity filter. To apply the diversity filter, we keep track of the number of times each medical paper appears in our corpus. Every time a new training example of a concept is added, we select the segment from the paper that appears the least frequently in the existing training data.

\subsection{Model}
%We use two separate transformer encoders for documents and concepts (Figure \ref{Figure:model}). 
Following the bi-encoder architecture~\cite {karpukhin2020dense}, documents and concepts are encoded separately by two encoders based on BERT~\cite{devlin2019bert}.
Specifically, the final hidden layer of the [CLS] token is used to represent document embeddings, while the concept encoder takes concatenated concept names and descriptions from the UMLS as input. %and similarly, we use the final layer representation of the [CLS] token as the concept embedding. 
%Given the extensive size of our concept collection, it is impractical to learn each concept embedding during every training step. Therefore, 
We pre-compute and freeze all concept embeddings before the training and inference phases. This strategy allows us to focus on training document embeddings to align accurately with concept embeddings.

% \begin{figure*}[!t]
% \centering
% \includegraphics[width=1\linewidth]{model.pdf}
%     \caption{Proposed Model Architecture for Document-level Concept Extraction}
% \label{Figure:model}
% \vspace{-2ex}
% \end{figure*}

This work examines 4 top-ranked models on the BLURB leaderboard \cite{10.1145/3458754} as the document and concept encoders: BioBERT \cite{DBLP:journals/corr/abs-1901-08746}, BiomedBERT \cite{10.1145/3458754}, SapBERT \cite{DBLP:journals/corr/abs-2010-11784}, and BioLinkBERT \cite{yasunaga-etal-2022-linkbert}. These models all share the BERT \citep{devlin-etal-2019-bert} architecture. BioBERT initializes its parameters and vocabulary from BERT and is pre-trained to do the same tasks as BERT on PMC \citep{PMC} and PubMed \citep{PubMed}. BiomedBERT is different from BioBERT because it trains its initial weights from scratch and uses a specialized medical vocabulary. SapBERT initializes its weights from BiomedBERT and learns to link entity mentions to entity IDs. BioLinkBERT initializes its weight from BiomedBERT and learns the citation links among medical documents.

\subsection{Training}
%After obtaining the document embedding from the encoder, we use the approximate vector search tool FAISS \cite{johnson2019billion} to select the best matching candidate concepts based on L2-distance. This process reportedly yields the top $k$ candidate concepts for each document 8.5 times faster than exact vector search. 
%In order to make the model learn the document embeddings, 
A common practice to train a semantic match model is based on contrastive learning with corresponding samples~\cite{karpukhin2020dense}.
We construct sets of positive and negative document-concept pairs. The positive set consists of documents along with their ground truth annotated concepts. The negative set combines random and hard negatives. We pick the irrelevant concepts that are among the top $k$ candidate concepts as the hard negatives, whereas the random negatives are drawn from the set of concepts which are neither positives nor hard negatives. Both types of negative samples are constructed dynamically during each training epoch to enhance the robustness of the model. %It is possible to have different negative document-concept pairs in different epochs given the same document, due to the variation of the document embedding in each epoch.

To effectively distinguish positive document-entity pairs from all the negative ones, we employ InfoNCE \citep{DBLP:journals/corr/abs-1807-03748} loss function weighted by the confidence of annotation for training. We first compute the embedding similarity of positive and negative document-entity pairs:
\begin{equation}
    E_p = e^{\text{sim}(f(\mathcal{D}), e_{p})}, \quad E_N = \sum_{i=1}^{N} e^{\text{sim}(f(\mathcal{D}), e_{n}^{i})}
\end{equation}
where sim denotes the cosine similarity function and $f(\mathcal{D})$, $e_{p}$, and $e_{n}$ denote the embedding of the document, positive concept, and negative concepts, respectively. Then we compute the InfoNCE loss of each training example as Eq.~\ref{eq: InfoNCE} and sum up the weighted InfoNCE of a batch as Eq.~\ref{eq: InfoNCE_batch}. 
\begin{equation}
\label{eq: InfoNCE}
    \mathcal{L} = - \frac{1}{P} \cdot \sum_{p=1}^{P} \left[ \log \frac{ E_p}{E_p + E_N} \right]
\end{equation} %and $N$ is the number of negative concepts of a document. 
\begin{equation}
\label{eq: InfoNCE_batch}
\mathcal{L} = w_a \sum_{a=1}^{A} \mathcal{L}_a + \sum_{m=1}^{M} \mathcal{L}_m
\end{equation}
where $P$ is the number of positive concepts of a document,  $\mathcal{L}_m$ is the InfoNCE of the $M$ manually annotated examples, and $\mathcal{L}_a$ is the InfoNCE of the $A$ augmented documents in a batch. Different from vanilla contrastive learning, our training corpus consists of manual and pseudo annotations generated by MetaMapLite. Since the MetaMapLite annotations are not entirely accurate, to capture the flaws of MetaMapLite annotations, a hyperparameter $w_a$ is adapted to assign a weight to each augmented example. 

\section{Experimental Setups}

\subsection{Target Data Characteristics}
Our evaluation datasets are BC5CDR \cite{li2016biocreative} and NCBI-Disease \cite{dougan2014ncbi}, whose statistic information is presented in Appendix~\ref{Sec:data_char}. There are plenty of non-canonical concept names and undertrained concepts in the corpus, which makes it challenging to predict.
%\color{blue}{(simply describe the focus (e.g., why suitable to use) and the difference between these two datasets and move the details to the appendix.)}
%\color{black}

\subsection{Data Preprocessing}
This work uses the standard tokenization and padding-to-maximum-length approach described by \citet{10.1145/3458754}. For the two-step pipeline experiments, we follow \cite{yasunaga-etal-2022-linkbert} to assign the word-level BIO labels to the sub-word level. Only the label of the initial token of a word contributes to the calculation of training loss and evaluation metrics. For document-level concept extraction, the label set is confined to the collection of unique labels in each target dataset.
% MeSH and OMIM both have well over 25,000 unique concepts. Due to our limited computational resources, we are unable to utilize the entire set of MeSH or OMIM ontologies for our concept label set. Instead, we construct the concept label set based on the target dataset. Specifically, only the concept IDs that appear in NCBI-Disease and BC5CDR are included in our label set. This brings the total size of our label set to under 3000. To ensure fairness in comparative experiments, we use the same concept label set for all our baselines and data-augmented approaches. The concept text for these IDs is sourced from MeSH and OMIM, as described in Section \ref{concept_text}. Subsequently, we pre-compute the concept embeddings using transformer models.

\subsection{Evaluation Metrics}
% We need to develop evaluation metrics that precisely measure the system's ability to extract unique concepts from documents. 
Token-level F1-score is commonly used for NER and EL.
%Traditional approaches often rely on token-level F1 scores for Named Entity Recognition (NER) and mention-level accuracy for Entity Linking (EL). 
However, these metrics fail to capture the true effectiveness of concept extraction, thus we employ document-level metrics for evaluation. %To provide enough candidates for evaluation, 
The top-10 concept predictions are selected for evaluation based on the
%approximate L2-distance 
similarity between the document and concept embeddings. We report the precision, recall, and F1-score for the top-10 candidate concepts. 
% We believe that ten concepts are sufficient to capture the major themes of each document, providing flexibility for post-processing if needed. We recognize that the number of unique concepts per document varies, ranging from 1 to 12 in NCBI-Disease and from 1 to 22 in BC5CDR. Any predicted concept within the top 10 that matches a correct concept is considered a true positive. Conversely, predicted concepts in the top 10 that do not match the gold standard are false positives, while gold standard concepts missing from the top 10 predictions are false negatives. 

\begin{table*}
\centering
%\scriptsize % Further reduce font size
%\setlength{\tabcolsep}{5pt} % Further adjust column spacing
\scalebox{0.75}{
\begin{tabular}{lcccccccccccc}
\toprule
      \multirow{3}{*}{{Method}} &  \multicolumn{6}{c}{NCBI-Disease} &  \multicolumn{6}{c}{BC5CDR}  \\
      \cmidrule(lr){2-7} \cmidrule(lr){8-13}
      &  \multicolumn{3}{c}{All} & \multicolumn{2}{c}{non-canonical} & Rare &  \multicolumn{3}{c}{All} & \multicolumn{2}{c}{non-canonical} & Rare  \\
      \cmidrule(lr){2-4} \cmidrule(lr){5-6} \cmidrule(lr){7-7} \cmidrule(lr){8-10} \cmidrule(lr){11-12} \cmidrule(lr){13-13}
      &  P & R & F1 & R@5 & R@10 & F1@10 & P & R & F1 & R@5 & R@10 & F1@10 \\
      \midrule
      BioLinkBERT-BioFEG & 0.4 & 1.2 & 0.6 & 2.5 & 2.8 & 2.2 & 0.2 & 0.3 & 0.2 & 0.2 & 0.5 & 0.1 \\
      MetaMapLite & 14.1 & 61.9 & 25.4 & 26.4  & 22.9 & 26.9 & 31.3 & 71.2 & 43.5 &  24.6 & 28.7 & 44.7 \\
      GPT-4 & 5.9 & 22.3 & 9.3 & 6.6 & 6.6 & 10.0 & 29.7 & 62.0 & 40.1 & 3.8 & 4.6 & 40.8 \\
     BioBERT & 16.8 & 80.7 & 27.9 & 34.4 & 69.9 & 30.2 & 24.1 & 58.9 & 34.2 & 21.3 & 46.0 & 34.0 \\
    BiomedBERT & 17.9 & 86.0 & 29.6 & 48.2 & 77.6 & 33.0 & 25.3 & 63.3 & 36.1 & 23.0 & 46.3 & 36.2 \\
    \hspace{2mm} + MMLite Aug. (Ours) &  20.4 & 92.6  & 33.5 & \textbf{48.3} & 82.2 & 39.2 & 34.8  & 82.2  & 48.9 & 27.8 & 57.6 & 49.8 \\
    \hspace{2mm} + SapBERT Aug. & 19.5 & 93.6 & 32.3 & 43.9 & 85.3 & 37.4 & 28.9 & 74.1 & 41.6 & 26.7 & 54.1 & 42.0 \\
    \hspace{2mm} + BioLinkBERT Aug. & 17.2 & 81.8 & 28.4 & 39.7 & 65.1 & 31.1 & 24.7 & 62.2 & 35.4 & 26.0 & 51.0 & 35.3 \\
        \midrule
    %   BioBERT & 16.8 & 80.7 & 27.9 & 34.4 & 69.9 & 30.2 & 24.1 & 58.9 & 34.2 & 21.3 & 46.0 & 34.0 \\
    %   \hspace{5mm} + MMLite Aug. (Ours) & 20.0 & 91.7 & 32.8 & 40.9 & 77.4 & 38.8 & 24.1 & 58.9 & 34.2 & 24.7 & 49.4 & 49.7 \\
    % SapBERT Aug. & 19.5 & 93.6 & 32.3 & 43.9 & 85.3 & 37.4 & 28.9 & 74.1 & 41.6 & 26.7 & 54.1 & 42.0 \\
     \hspace{2mm} + SapBERT w/ MMLite (Ours) & \textbf{21.1} & \textbf{95.3} & \textbf{34.5} & 47.6 & \textbf{87.3} & \textbf{41.1} & \textbf{35.8} & \textbf{84.5} & \textbf{50.3} & \textbf{30.7} & \textbf{62.0} & 51.4 \\
      % BioLinkBERT Aug. & 17.2 & 81.8 & 28.4 & 39.7 & 65.1 & 31.1 & 24.7 & 62.2 & 35.4 & 26.0 & 51.0 & 35.3 \\
      \hspace{2mm} + BioLinkBERT w/ MMLite (Ours) & 20.2 & 91.1 & 33.0 & 44.5 & 76.0 & 38.4 & 35.4 & 84.1 & 49.8 & 26.4 & 53.8 & \textbf{51.8} \\
\bottomrule
\end{tabular}}
\caption{\label{Table:data_aug_comparison} Performance of concept extraction results on different systems. \textbf{Bold} indicates the best performance. ``All'', ``non-canonical'', and ``Rare'' denote performance on all, non-canonical and rare test concepts.}
\vspace{-2ex}
\end{table*}

\subsection{Baseline and Implementation}
We select the state-of-the-art models featured on the BLURB leaderboard\footnote{\url{https://microsoft.github.io/BLURB/tasks.html}} as baseline methods, including BioBERT \citep{DBLP:journals/corr/abs-1901-08746}, BiomedBERT \cite{10.1145/3458754}, SapBERT \cite{DBLP:journals/corr/abs-2010-11784}, and BioLinkBERT \cite{yasunaga-etal-2022-linkbert}. 
For fair comparison, all models use the standard hyper-parameter configurations of these models when they are fine-tuned either with the original manually annotated data or with the augmented data. 
During training, each positive candidate is contrasted with 25 negative candidates: 5 random negative candidates drawn from a uniform distribution of concept IDs, and 20 hard negative candidates, which are the 20 closest negative candidates to the document in the embedding space. For the GPT-4 baseline, we use the prompt template for each document: \textit{``Given the concept collection [Concept Collection], identify 10 concepts mentioned in the document [Document]''}. The MetaMapLite \cite{demner2017metamap} and the two-step baseline with BioLinkBERT and BioEFG are implemented using the hyper-parameters of the original work.

% \subsection{Implementation Details}
% Our model utilizes a standard hyperparameter setup of transformer architectures. During training, we contrast each positive candidate with 20 hard negative candidates and 5 random negative candidates. The 20 hard negative candidates are the closest negative candidates to the document in the embedding space. For inference, we select only 10 candidate concepts for evaluation. 

% For the GPT-4 baseline, we use the OpenAI Chat API with the GPT-4 model. We construct a standardized message template for each document: \textit{"Given the concept collection [Concept Collection], identify 10 concepts mentioned in the document [Document]."} To ensure consistency with the transformer models, we incorporate all unique concepts found in the dataset into the concept collection.

% We employ the Java implementation of MetaMapLite to annotate all the documents. To ensure a fair comparison, we post-process the MetaMapLite predictions by converting the UMLS concept IDs to MeSH or OMIM concept IDs and removing any predicted concepts that are not present in the concept collection of the target dataset.

% For the NER+EL approach, we use BioLinkBERT for NER and BioFEG for EL. We select the concept mention spans predicted by BioLinkBERT as input for BioFEG. Subsequently, we take the top concept ID for each mention span predicted by BioFEG and remove any duplicate concept IDs within a document. This process formulates the predicted concepts on the document-level.

\section{Results and Analysis}
\subsection{Main Results} \label{Sec:main_exp}
\textbf{Overall Performance} \quad The effectiveness of our data-augmented training strategy is demonstrated in the columns under ``All'' in Table~\ref{Table:data_aug_comparison}. Compared to training on manually annotated data alone, our method outperforms all of the baselines because the baselines lack training data for rare and non-canonical concepts. Additionally, our data augmentation approach surpasses other techniques like SapBERT and BioLinkBERT when initialized from BiomedBERT. To test if our approach is complementary to %can also integrate with 
other augmentation techniques, we combine our augmentation with SapBERT and BioLinkBERT augmentation approaches. % to further enhance results, addressing the lack of training examples for rare concepts in the original corpus, as shown in the columns under ``All'' in Table \ref{Table:data_aug_comparison}.
The results in Table~\ref{Table:data_aug_comparison} show that combining with SapBERT further increases the performance while combining with BioLinkBERT does not. Our augmentation approach and SapBERT are thus complementary to some extent.

\noindent \textbf{Performance on Rare Concepts} \quad Different from \citet{DBLP:journals/corr/abs-1901-08746}, \citet{kanakarajan-etal-2021-bioelectra}, \citet{10.1145/3458754}, \citet{DBLP:journals/corr/abs-2010-11784}, and \citet{yasunaga-etal-2022-linkbert} that transfer embeddings from loosely related tasks, our technique directly targets low-resource concepts in concept extraction tasks. We show the performance of models on concepts with fewer than 10 manually annotated training examples in the columns under ``Rare'' of Table \ref{Table:data_aug_comparison}. The results confirm that our data augmentation method significantly improves the F1 score for concepts with insufficient manual annotations, explaining its superior performance over alternative methods.

\noindent \textbf{Performance on Non-canonical Concepts} \quad One of the key challenges for existing models is non-canonical concept prediction. This task is difficult because non-canonical concept mentions often do not match the canonical names of concepts stored in knowledge graphs. 
% Finding training data that directly links non-canonical concept mentions to concept IDs is particularly challenging. 
We have observed that non-canonical concept mentions often share similar contexts with canonical concept names. Our proposed method enables models to infer concepts from a variety of contexts. The columns under ``non-canonical'' in Table \ref{Table:data_aug_comparison} demonstrate that models trained with augmented data outperform those trained exclusively on manually annotated data in predicting non-canonical concepts. 
% These models also surpass the performance of the two-step approach using BioLinkBERT and BioFEG, MetaMapLite, and GPT-4. 
This indicates that learning from diverse, even noisy, contexts can enhance the prediction of non-canonical concepts.

\subsection{Quantifying the Effect of Data Augmentation} \label{Sec:data_quant}
\begin{figure}[!t]
\centering
\includegraphics[width=1\linewidth]{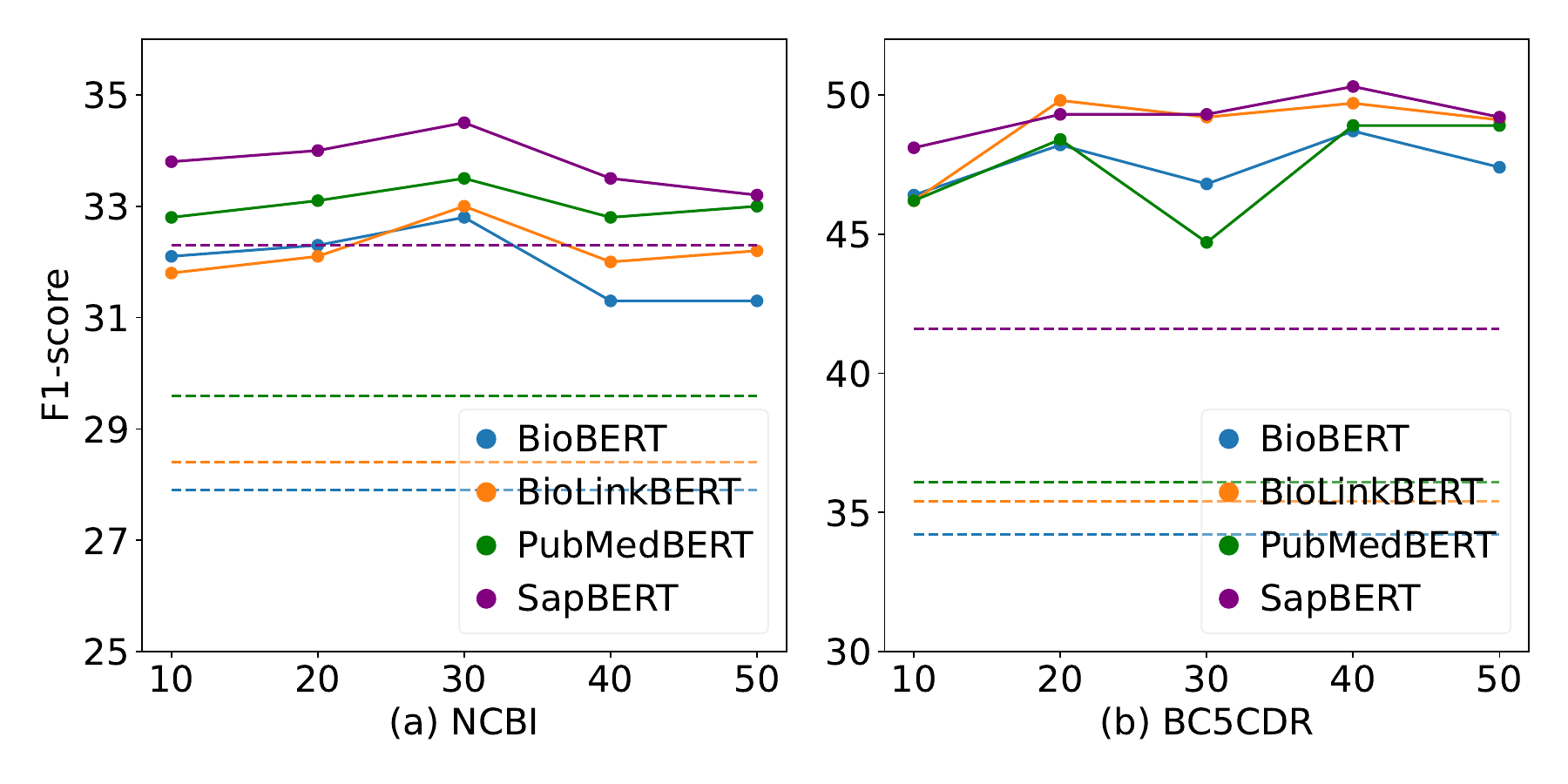}
    \caption{F1-score with different entity occurrence threshold $k$ on NCBI-Disease (a) and BC5CDR (b). The solid curves show the F1-scores with different augmentation quantities, and the dotted lines show the baseline F1-scores without augmentation.}
\label{freq_f1}
\vspace{-2ex}
\end{figure}

To quantify how the training occurrence threshold $k$ impacts the performance of our data augmentation method, we test 4 data-augmented models at different values of $k$.
Figure \ref{freq_f1} shows a big performance boost when going from no augmented data to a small quantity of augmented data, but the effect plateaus once $k$ is above 10. This might be mainly because a $k$ value above 10 makes the model learn from a variety of documents where a concept is paraphrased in multiple ways, which makes the training process more robust to variations of concepts.
% BiomedBERT is more sensitive to $k$ than other models. 
Both the validation and test sets exhibit similar degrees of performance fluctuation, where the F1 score peaks at the same level of $k$. This consistency enables us to select the optimal $k$ value based on validation set performance.

\begin{table}
    \centering
    \scalebox{0.85}{
    \begin{tabular}{lcc}
        \toprule
        \multirow{2}{*}{{Method}} & \multicolumn{2}{c}{BC5CDR}  \\
        \cmidrule(lr){2-3} 
        ~ & Chemical & Disease\\
        \midrule
        MetaMapLite & 38.8 & 35.1 \\
        BioBERT & 28.5 & 31.6  \\
        % \hspace{5mm} + MMLite Aug. (Ours) & 41.8 & 40.6 \\
        BiomedBERT & 31.2 & 31.7 \\
        \hspace{5mm} + MMLite Aug. (Ours) & \textbf{39.1} & \textbf{36.5} \\
        \hspace{5mm} + BioLinkBERT & 29.5 & 31.1 \\
        % \hspace{10mm} + MMLite Aug. (Ours) & 42.0 & 41.5 \\
        \hspace{5mm} + SapBERT & 35.5 & 35.1 \\
        % \hspace{10mm} + MMLite Aug. (Ours) & \textbf{44.3} & \textbf{42.9} \\
        \bottomrule
    \end{tabular}}
    \caption{\label{Table:concept_type}
    F1-score on chemical and disease concepts of BC5CDR. \textbf{Bold} indicates the best performance.}
\end{table}

% \begin{table*}
%     \centering
%     \scalebox{0.75}{%
%         \begin{tabular}{lcccccccc}
%             \toprule
%             \multirow{2}{*}{{Method}} & \multicolumn{2}{c}{BioBERT} & \multicolumn{2}{c}{BiomedBERT} & \multicolumn{2}{c}{BioLinkBERT} & \multicolumn{2}{c}{SapBERT}\\
%             \cmidrule(lr){2-3} \cmidrule(lr){4-5} \cmidrule(lr){6-7} \cmidrule(lr){8-9}
%             & NCBI-Disease & BC5CDR & NCBI-Disease & BC5CDR & NCBI-Disease & BC5CDR & NCBI-Disease & BC5CDR\\
%             \midrule
%             No Filter & 31.5 & 45.8 & 33.7 & 48.8 & 32.2 & 48.0 & 32.7 & 45.2 \\
%             Abv. Filter & 32.0 & 45.9 & 33.8 & 49.1 & 32.3 & 48.3 & 33.1 & 45.8 \\
%             Olap. Filter & 32.8 & 48.4 & 34.4 & 50.2  & 32.8 & 49.4 & 33.3 & 47.6 \\
%             Div. Filter & 32.2 & 46.4 & 34.1 & 50.0 & 32.6 & 49.4 & 33.2 & 46.7 \\
%             All Filters & 32.9 & 48.6 & 34.5 & 50.3 & 33.0 & 49.7 & 33.5 & 48.9\\
%             \bottomrule
%         \end{tabular}%
%     }
%     \caption{\label{Table:filter_abl} F1-score of Models with Filters}
% \end{table*}

\subsection{Modeling the Quality of Pseudo-annotations} \label{Sec:aug_loss}
\begin{figure}[!t]
\centering
\includegraphics[width=1\linewidth]{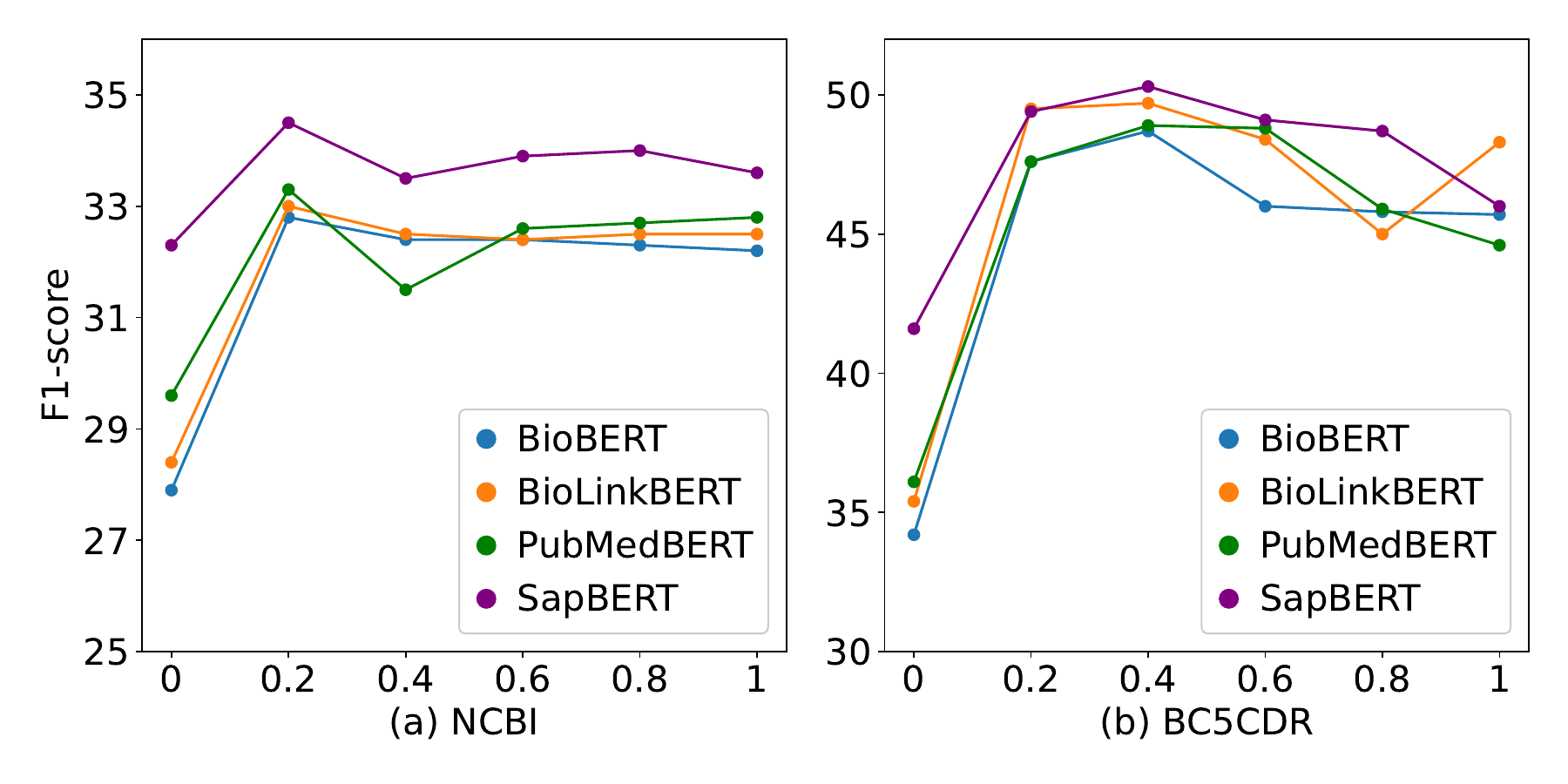}
    \caption{F1 with different augmented weight on NCBI-Disease (a) and BC5CDR (b).}
\label{agwt_f1}
\vspace{-2ex}
\end{figure}

We capture the quality of pseudo-annotations by MetaMapLite using the augmented weight $w_a$. Figure \ref{agwt_f1} shows the effect of $w_a$ on the F1 for BC5CDR and NCBI-Disease. The optimal augmented weight for the models falls around 0.4 on BC5CDR and 0.2 on NCBI-Disease, which reflects the quality of pseudo-annotations. Different values of  $w_a$ can cause a fluctuation in F1 of roughly 10\%. The impact of augmented weight is similar between the test and validation set, which allows us to use the validation set to guide our choice of augmented weight.

\subsection{Impact on Different Concept Types} \label{Sec:c_type}
Data augmentation affects chemical and disease concepts differently. We quantify the disparity in Table \ref{Table:concept_type}, which could stem from the higher error rate of our pseudo-annotation tool, MetaMapLite, when identifying disease concepts. Disease concepts are particularly challenging for MetaMapLite due to the subtle distinctions between narrow and broad concepts. For example, MetaMapLite often mistakenly equates kidney disease with kidney damage, although they are distinct concepts according to UMLS. In contrast, chemical concepts are usually written as exact chemical formulations in our training corpus, making them easier for MetaMapLite to identify accurately. This difference in annotation quality between chemical and disease concepts is reflected in the performance of data-augmented models.

\subsection{Effect of Filters} \label{Sec:filters}
\begin{figure}[!t]
\centering
\includegraphics[width=1\linewidth]{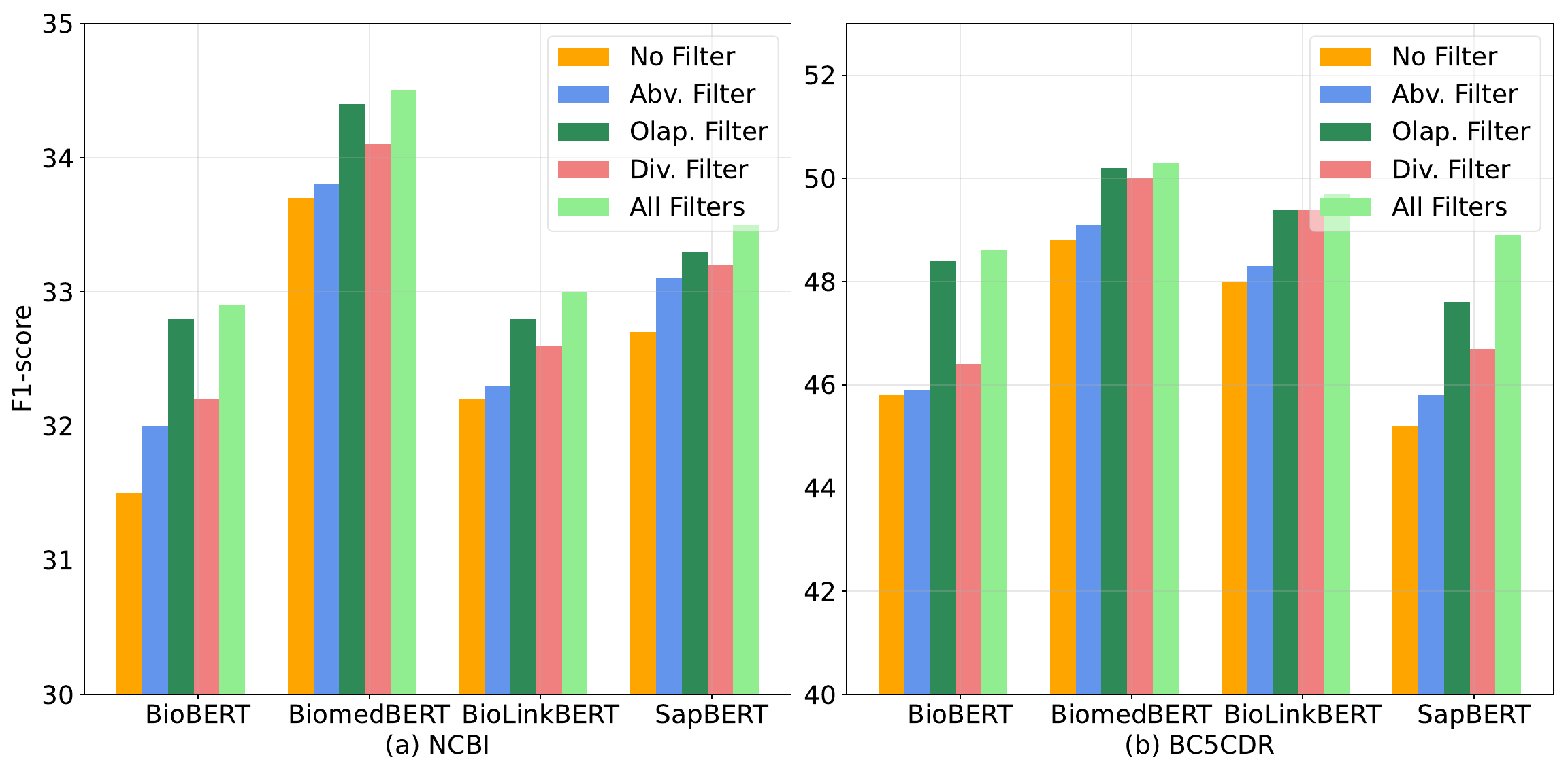}
    \caption{F1-score of models with different filters on NCBI (a) and BC5CDR (b).}
\label{Figure:filters}
\vspace{-2ex}
\end{figure}

To examine the effect of the post-annotation filters, we conduct an ablation study by adding one filter at a time.
According to Figure \ref{Figure:filters}, all 3 filters improve the F1 score. The biggest improvement comes from the overlap filter. This is because the overlap filter is consistent with the practice of manual annotation. The abbreviation filter does not have a huge effect because it only affects very few annotations within our corpus. The diversity filter does not have a huge effect either, because most concepts within the test data are extracted from one single document. Nonetheless, % we notice it is a good practice to combine all three filters because 
the combination of all filters performs the best on our test data. %, and can be generalized to use cases with a greater variety of concepts and documents.

\section{Conclusion}
 In this study, we address the challenge of data scarcity and non-canonical concept names in document-level concept extraction via a novel pseudo-annotation approach with MetaMapLite for data augmentation. Our methodology facilitates a contextualized process, significantly improving model performance on benchmark datasets, especially under conditions of limited labeled training data. Our experiments not only validate the proposed approach but also provide valuable insights for optimizing data augmentation strategies in biomedical text mining, marking a significant step forward in the extraction and retrieval of biomedical information.

\section*{Limitations}
Given the scarcity of research on document-level concept extraction, we made significant efforts to gather various open-source baselines with good reported performance on related tasks. We implemented the concept extraction experiments using these baselines all by ourselves. It is possible that some baseline models may require specific hyperparameter configurations for this task beyond those reported in the literature. We leave the detailed hyperparameter tuning and related investigations for future work.

%\section*{Ethics Statement}
%Scientific work published at ACL 2023 must comply with the ACL Ethics Policy.\footnote{\url{https://www.aclweb.org/portal/content/acl-code-ethics}} We encourage all authors to include an explicit ethics statement on the broader impact of the work, or other ethical considerations after the conclusion but before the references. The ethics statement will not count toward the page limit (8 pages for long, 4 pages for short papers).

%\section*{Acknowledgements}

\bibliography{anthology,custom}

\begin{thebibliography}{48}
\expandafter\ifx\csname natexlab\endcsname\relax\def\natexlab#1{#1}\fi

\bibitem[{Alrowili and Shanker(2021)}]{alrowili-shanker-2021-biom}
Sultan Alrowili and Vijay Shanker. 2021.
\newblock \href {https://doi.org/10.18653/v1/2021.bionlp-1.24} {{B}io{M}-transformers: Building large biomedical language models with {BERT}, {ALBERT} and {ELECTRA}}.
\newblock In \emph{Proceedings of the 20th Workshop on Biomedical Language Processing}, pages 221--227, Online. Association for Computational Linguistics.

\bibitem[{Angell et~al.(2021)Angell, Monath, Mohan, Yadav, and McCallum}]{angell-etal-2021-clustering}
Rico Angell, Nicholas Monath, Sunil Mohan, Nishant Yadav, and Andrew McCallum. 2021.
\newblock \href {https://doi.org/10.18653/v1/2021.naacl-main.205} {Clustering-based inference for biomedical entity linking}.
\newblock In \emph{Proceedings of the 2021 Conference of the North American Chapter of the Association for Computational Linguistics: Human Language Technologies}, pages 2598--2608, Online. Association for Computational Linguistics.

\bibitem[{Aronson and Lang(2010)}]{aronson2010overview}
Alan~R Aronson and Fran{\c{c}}ois-Michel Lang. 2010.
\newblock An overview of metamap: historical perspective and recent advances.
\newblock \emph{Journal of the American Medical Informatics Association}, 17(3):229--236.

\bibitem[{Bartolini et~al.(2023)Bartolini, Moscato, Postiglione, Sperlì, and Vignali}]{BARTOLINI2023102291}
Ilaria Bartolini, Vincenzo Moscato, Marco Postiglione, Giancarlo Sperlì, and Andrea Vignali. 2023.
\newblock \href {https://doi.org/https://doi.org/10.1016/j.is.2023.102291} {Data augmentation via context similarity: An application to biomedical named entity recognition}.
\newblock \emph{Information Systems}, 119:102291.

\bibitem[{Beltagy et~al.(2019)Beltagy, Lo, and Cohan}]{beltagy-etal-2019-scibert}
Iz~Beltagy, Kyle Lo, and Arman Cohan. 2019.
\newblock \href {https://doi.org/10.18653/v1/D19-1371} {{S}ci{BERT}: A pretrained language model for scientific text}.
\newblock In \emph{Proceedings of the 2019 Conference on Empirical Methods in Natural Language Processing and the 9th International Joint Conference on Natural Language Processing (EMNLP-IJCNLP)}, pages 3615--3620, Hong Kong, China. Association for Computational Linguistics.

\bibitem[{Bhowmik et~al.(2021)Bhowmik, Stratos, and de~Melo}]{bhowmik-etal-2021-fast}
Rajarshi Bhowmik, Karl Stratos, and Gerard de~Melo. 2021.
\newblock \href {https://aclanthology.org/2021.louhi-1.4} {Fast and effective biomedical entity linking using a dual encoder}.
\newblock In \emph{Proceedings of the 12th International Workshop on Health Text Mining and Information Analysis}, pages 28--37, online. Association for Computational Linguistics.

\bibitem[{Bodenreider(2004)}]{bodenreider2004unified}
Olivier Bodenreider. 2004.
\newblock The unified medical language system (umls): integrating biomedical terminology.
\newblock \emph{Nucleic acids research}, 32(suppl\_1):D267--D270.

\bibitem[{Demner-Fushman et~al.(2017)Demner-Fushman, Rogers, and Aronson}]{demner2017metamap}
Dina Demner-Fushman, Willie~J Rogers, and Alan~R Aronson. 2017.
\newblock Metamap lite: an evaluation of a new java implementation of metamap.
\newblock \emph{Journal of the American Medical Informatics Association}, 24(4):841--844.

\bibitem[{Devlin et~al.(2019{\natexlab{a}})Devlin, Chang, Lee, and Toutanova}]{devlin2019bert}
Jacob Devlin, Ming-Wei Chang, Kenton Lee, and Kristina Toutanova. 2019{\natexlab{a}}.
\newblock Bert: Pre-training of deep bidirectional transformers for language understanding.
\newblock In \emph{Proceedings of NAACL-HLT}, pages 4171--4186.

\bibitem[{Devlin et~al.(2019{\natexlab{b}})Devlin, Chang, Lee, and Toutanova}]{devlin-etal-2019-bert}
Jacob Devlin, Ming-Wei Chang, Kenton Lee, and Kristina Toutanova. 2019{\natexlab{b}}.
\newblock \href {https://doi.org/10.18653/v1/N19-1423} {{BERT}: Pre-training of deep bidirectional transformers for language understanding}.
\newblock In \emph{Proceedings of the 2019 Conference of the North {A}merican Chapter of the Association for Computational Linguistics: Human Language Technologies, Volume 1 (Long and Short Papers)}, pages 4171--4186, Minneapolis, Minnesota. Association for Computational Linguistics.

\bibitem[{Do{\u{g}}an et~al.(2014)Do{\u{g}}an, Leaman, and Lu}]{dougan2014ncbi}
Rezarta~Islamaj Do{\u{g}}an, Robert Leaman, and Zhiyong Lu. 2014.
\newblock Ncbi disease corpus: a resource for disease name recognition and concept normalization.
\newblock \emph{Journal of biomedical informatics}, 47:1--10.

\bibitem[{Fei et~al.(2021)Fei, Ren, Zhang, Ji, and Liang}]{fei2021enriching}
Hao Fei, Yafeng Ren, Yue Zhang, Donghong Ji, and Xiaohui Liang. 2021.
\newblock Enriching contextualized language model from knowledge graph for biomedical information extraction.
\newblock \emph{Briefings in bioinformatics}, 22(3):bbaa110.

\bibitem[{Gu et~al.(2020)Gu, Tinn, Cheng, Lucas, Usuyama, Liu, Naumann, Gao, and Poon}]{DBLP:journals/corr/abs-2007-15779}
Yu~Gu, Robert Tinn, Hao Cheng, Michael Lucas, Naoto Usuyama, Xiaodong Liu, Tristan Naumann, Jianfeng Gao, and Hoifung Poon. 2020.
\newblock \href {http://arxiv.org/abs/2007.15779} {Domain-specific language model pretraining for biomedical natural language processing}.
\newblock \emph{CoRR}, abs/2007.15779.

\bibitem[{Gu et~al.(2021)Gu, Tinn, Cheng, Lucas, Usuyama, Liu, Naumann, Gao, and Poon}]{10.1145/3458754}
Yu~Gu, Robert Tinn, Hao Cheng, Michael Lucas, Naoto Usuyama, Xiaodong Liu, Tristan Naumann, Jianfeng Gao, and Hoifung Poon. 2021.
\newblock \href {https://doi.org/10.1145/3458754} {Domain-specific language model pretraining for biomedical natural language processing}.
\newblock \emph{ACM Trans. Comput. Healthcare}, 3(1).

\bibitem[{Johnson et~al.(2019)Johnson, Douze, and J{\'e}gou}]{johnson2019billion}
Jeff Johnson, Matthijs Douze, and Herv{\'e} J{\'e}gou. 2019.
\newblock Billion-scale similarity search with {GPUs}.
\newblock \emph{IEEE Transactions on Big Data}, 7(3):535--547.

\bibitem[{Kanakarajan et~al.(2021)Kanakarajan, Kundumani, and Sankarasubbu}]{kanakarajan-etal-2021-bioelectra}
Kamal~raj Kanakarajan, Bhuvana Kundumani, and Malaikannan Sankarasubbu. 2021.
\newblock \href {https://doi.org/10.18653/v1/2021.bionlp-1.16} {{B}io{ELECTRA}:pretrained biomedical text encoder using discriminators}.
\newblock In \emph{Proceedings of the 20th Workshop on Biomedical Language Processing}, pages 143--154, Online. Association for Computational Linguistics.

\bibitem[{Karpukhin et~al.(2020)Karpukhin, Oguz, Min, Lewis, Wu, Edunov, Chen, and Yih}]{karpukhin2020dense}
Vladimir Karpukhin, Barlas Oguz, Sewon Min, Patrick Lewis, Ledell Wu, Sergey Edunov, Danqi Chen, and Wen-tau Yih. 2020.
\newblock Dense passage retrieval for open-domain question answering.
\newblock In \emph{Proceedings of the 2020 Conference on Empirical Methods in Natural Language Processing (EMNLP)}, pages 6769--6781.

\bibitem[{Lee et~al.(2019)Lee, Yoon, Kim, Kim, Kim, So, and Kang}]{DBLP:journals/corr/abs-1901-08746}
Jinhyuk Lee, Wonjin Yoon, Sungdong Kim, Donghyeon Kim, Sunkyu Kim, Chan~Ho So, and Jaewoo Kang. 2019.
\newblock \href {http://arxiv.org/abs/1901.08746} {Biobert: a pre-trained biomedical language representation model for biomedical text mining}.
\newblock \emph{CoRR}, abs/1901.08746.

\bibitem[{Li et~al.(2019)Li, Jin, Liu, Rawat, Cai, and Yu}]{info:doi/10.2196/14830}
Fei Li, Yonghao Jin, Weisong Liu, Bhanu Pratap~Singh Rawat, Pengshan Cai, and Hong Yu. 2019.
\newblock \href {https://doi.org/10.2196/14830} {Fine-tuning bidirectional encoder representations from transformers (bert)--based models on large-scale electronic health record notes: An empirical study}.
\newblock \emph{JMIR Med Inform}, 7(3):e14830.

\bibitem[{Li et~al.(2021)Li, Ding, Shang, McAuley, and Feng}]{li2021weakly}
Jiacheng Li, Haibo Ding, Jingbo Shang, Julian McAuley, and Zhe Feng. 2021.
\newblock \href {http://arxiv.org/abs/2107.02282} {Weakly supervised named entity tagging with learnable logical rules}.

\bibitem[{Li et~al.(2016)Li, Sun, Johnson, Sciaky, Wei, Leaman, Davis, Mattingly, Wiegers, and Lu}]{li2016biocreative}
Jiao Li, Yueping Sun, Robin~J Johnson, Daniela Sciaky, Chih-Hsuan Wei, Robert Leaman, Allan~Peter Davis, Carolyn~J Mattingly, Thomas~C Wiegers, and Zhiyong Lu. 2016.
\newblock Biocreative v cdr task corpus: a resource for chemical disease relation extraction.
\newblock \emph{Database}, 2016.

\bibitem[{Liu et~al.(2020{\natexlab{a}})Liu, Shareghi, Meng, Basaldella, and Collier}]{DBLP:journals/corr/abs-2010-11784}
Fangyu Liu, Ehsan Shareghi, Zaiqiao Meng, Marco Basaldella, and Nigel Collier. 2020{\natexlab{a}}.
\newblock \href {http://arxiv.org/abs/2010.11784} {Self-alignment pre-training for biomedical entity representations}.
\newblock \emph{CoRR}, abs/2010.11784.

\bibitem[{Liu et~al.(2020{\natexlab{b}})Liu, Shareghi, Meng, Basaldella, and Collier}]{liu2020self}
Fangyu Liu, Ehsan Shareghi, Zaiqiao Meng, Marco Basaldella, and Nigel Collier. 2020{\natexlab{b}}.
\newblock Self-alignment pretraining for biomedical entity representations.
\newblock \emph{arXiv preprint arXiv:2010.11784}.

\bibitem[{Liu et~al.(2021)Liu, Vuli{\'c}, Korhonen, and Collier}]{liu2021fast}
Fangyu Liu, Ivan Vuli{\'c}, Anna Korhonen, and Nigel Collier. 2021.
\newblock Fast, effective, and self-supervised: Transforming masked language models into universal lexical and sentence encoders.
\newblock \emph{arXiv preprint arXiv:2104.08027}.

\bibitem[{Liu et~al.(2020{\natexlab{c}})Liu, Sun, Li, Wang, and Zhao}]{Liu_Sun_Li_Wang_Zhao_2020}
Shifeng Liu, Yifang Sun, Bing Li, Wei Wang, and Xiang Zhao. 2020{\natexlab{c}}.
\newblock \href {https://doi.org/10.1609/aaai.v34i05.6358} {Hamner: Headword amplified multi-span distantly supervised method for domain specific named entity recognition}.
\newblock \emph{Proceedings of the AAAI Conference on Artificial Intelligence}, 34(05):8401--8408.

\bibitem[{Michalopoulos et~al.(2021)Michalopoulos, Wang, Kaka, Chen, and Wong}]{michalopoulos-etal-2021-umlsbert}
George Michalopoulos, Yuanxin Wang, Hussam Kaka, Helen Chen, and Alexander Wong. 2021.
\newblock \href {https://doi.org/10.18653/v1/2021.naacl-main.139} {{U}mls{BERT}: Clinical domain knowledge augmentation of contextual embeddings using the {U}nified {M}edical {L}anguage {S}ystem {M}etathesaurus}.
\newblock In \emph{Proceedings of the 2021 Conference of the North American Chapter of the Association for Computational Linguistics: Human Language Technologies}, pages 1744--1753, Online. Association for Computational Linguistics.

\bibitem[{Miftahutdinov et~al.(2021)Miftahutdinov, Kadurin, Kudrin, and Tutubalina}]{miftahutdinov2021medical}
Zulfat Miftahutdinov, Artur Kadurin, Roman Kudrin, and Elena Tutubalina. 2021.
\newblock Medical concept normalization in clinical trials with drug and disease representation learning.
\newblock \emph{Bioinformatics}, 37(21):3856--3864.

\bibitem[{Nentidis et~al.(2020)Nentidis, Bougiatiotis, Krithara, and Paliouras}]{nentidis2020results}
Anastasios Nentidis, Konstantinos Bougiatiotis, Anastasia Krithara, and Georgios Paliouras. 2020.
\newblock Results of the seventh edition of the bioasq challenge.
\newblock In \emph{Machine Learning and Knowledge Discovery in Databases: International Workshops of ECML PKDD 2019, W{\"u}rzburg, Germany, September 16--20, 2019, Proceedings, Part II}, pages 553--568. Springer.

\bibitem[{Neumann et~al.(2019)Neumann, King, Beltagy, and Ammar}]{neumann2019scispacy}
Mark Neumann, Daniel King, Iz~Beltagy, and Waleed Ammar. 2019.
\newblock Scispacy: fast and robust models for biomedical natural language processing.
\newblock \emph{arXiv preprint arXiv:1902.07669}.

\bibitem[{NIH(2023{\natexlab{a}})}]{PubMed}
NIH. 2023{\natexlab{a}}.
\newblock \href {https://pubmed.ncbi.nlm.nih.gov/} {Pubmed}.
\newblock Accessed: 2023-12-22.

\bibitem[{NIH(2023{\natexlab{b}})}]{PMC}
NIH. 2023{\natexlab{b}}.
\newblock \href {https://www.ncbi.nlm.nih.gov/pmc/} {Pubmed central}.
\newblock Accessed: 2023-12-22.

\bibitem[{Peng et~al.(2019)Peng, Yan, and Lu}]{peng-etal-2019-transfer}
Yifan Peng, Shankai Yan, and Zhiyong Lu. 2019.
\newblock \href {https://doi.org/10.18653/v1/W19-5006} {Transfer learning in biomedical natural language processing: An evaluation of {BERT} and {ELM}o on ten benchmarking datasets}.
\newblock In \emph{Proceedings of the 18th BioNLP Workshop and Shared Task}, pages 58--65, Florence, Italy. Association for Computational Linguistics.

\bibitem[{Phan et~al.(2021)Phan, Anibal, Tran, Chanana, Bahadroglu, Peltekian, and Altan{-}Bonnet}]{DBLP:journals/corr/abs-2106-03598}
Long~N. Phan, James~T. Anibal, Hieu Tran, Shaurya Chanana, Erol Bahadroglu, Alec Peltekian, and Gr{\'{e}}goire Altan{-}Bonnet. 2021.
\newblock \href {http://arxiv.org/abs/2106.03598} {Scifive: a text-to-text transformer model for biomedical literature}.
\newblock \emph{CoRR}, abs/2106.03598.

\bibitem[{Rohanian et~al.(2023)Rohanian, Nouriborji, Kouchaki, and Clifton}]{10.1093/bioinformatics/btad103}
Omid Rohanian, Mohammadmahdi Nouriborji, Samaneh Kouchaki, and David~A Clifton. 2023.
\newblock \href {https://doi.org/10.1093/bioinformatics/btad103} {{On the effectiveness of compact biomedical transformers}}.
\newblock \emph{Bioinformatics}, 39(3):btad103.

\bibitem[{Sachan et~al.(2018)Sachan, Xie, Sachan, and Xing}]{pmlr-v85-sachan18a}
Devendra~Singh Sachan, Pengtao Xie, Mrinmaya Sachan, and Eric~P. Xing. 2018.
\newblock \href {https://proceedings.mlr.press/v85/sachan18a.html} {Effective use of bidirectional language modeling for transfer learning in biomedical named entity recognition}.
\newblock In \emph{Proceedings of the 3rd Machine Learning for Healthcare Conference}, volume~85 of \emph{Proceedings of Machine Learning Research}, pages 383--402. PMLR.

\bibitem[{Savova et~al.(2010)Savova, Masanz, Ogren, Zheng, Sohn, Kipper-Schuler, and Chute}]{savova2010mayo}
Guergana~K Savova, James~J Masanz, Philip~V Ogren, Jiaping Zheng, Sunghwan Sohn, Karin~C Kipper-Schuler, and Christopher~G Chute. 2010.
\newblock Mayo clinical text analysis and knowledge extraction system (ctakes): architecture, component evaluation and applications.
\newblock \emph{Journal of the American Medical Informatics Association}, 17(5):507--513.

\bibitem[{Shin et~al.(2020)Shin, Zhang, Bakhturina, Puri, Patwary, Shoeybi, and Mani}]{shin-etal-2020-biomegatron}
Hoo-Chang Shin, Yang Zhang, Evelina Bakhturina, Raul Puri, Mostofa Patwary, Mohammad Shoeybi, and Raghav Mani. 2020.
\newblock \href {https://doi.org/10.18653/v1/2020.emnlp-main.379} {{B}io{M}egatron: Larger biomedical domain language model}.
\newblock In \emph{Proceedings of the 2020 Conference on Empirical Methods in Natural Language Processing (EMNLP)}, pages 4700--4706, Online. Association for Computational Linguistics.

\bibitem[{Soldaini and Goharian(2016)}]{soldaini2016quickumls}
Luca Soldaini and Nazli Goharian. 2016.
\newblock Quickumls: a fast, unsupervised approach for medical concept extraction.
\newblock In \emph{MedIR workshop, sigir}, pages 1--4.

\bibitem[{Sui et~al.(2023)Sui, Zhang, Cai, Song, Zhou, Yuan, and Zhang}]{sui2023biofeg}
Xuhui Sui, Ying Zhang, Xiangrui Cai, Kehui Song, Baohang Zhou, Xiaojie Yuan, and Wensheng Zhang. 2023.
\newblock Biofeg: Generate latent features for biomedical entity linking.
\newblock In \emph{Proceedings of the 2023 Conference on Empirical Methods in Natural Language Processing}, pages 11584--11593.

\bibitem[{Sung et~al.(2020)Sung, Jeon, Lee, and Kang}]{sung2020biomedical}
Mujeen Sung, Hwisang Jeon, Jinhyuk Lee, and Jaewoo Kang. 2020.
\newblock Biomedical entity representations with synonym marginalization.
\newblock \emph{arXiv preprint arXiv:2005.00239}.

\bibitem[{Tang et~al.(2023)Tang, Han, Jiang, and Hu}]{tang2023does}
Ruixiang Tang, Xiaotian Han, Xiaoqian Jiang, and Xia Hu. 2023.
\newblock \href {http://arxiv.org/abs/2303.04360} {Does synthetic data generation of llms help clinical text mining?}

\bibitem[{van~den Oord et~al.(2018)van~den Oord, Li, and Vinyals}]{DBLP:journals/corr/abs-1807-03748}
A{\"{a}}ron van~den Oord, Yazhe Li, and Oriol Vinyals. 2018.
\newblock \href {http://arxiv.org/abs/1807.03748} {Representation learning with contrastive predictive coding}.
\newblock \emph{CoRR}, abs/1807.03748.

\bibitem[{Varma et~al.(2021)Varma, Orr, Wu, Leszczynski, Ling, and R{\'e}}]{varma2021cross}
Maya Varma, Laurel Orr, Sen Wu, Megan Leszczynski, Xiao Ling, and Christopher R{\'e}. 2021.
\newblock Cross-domain data integration for named entity disambiguation in biomedical text.
\newblock \emph{arXiv preprint arXiv:2110.08228}.

\bibitem[{Wang et~al.(2022)Wang, Jiang, Bach, Wang, Huang, Huang, and Tu}]{wang2022improving}
Xinyu Wang, Yong Jiang, Nguyen Bach, Tao Wang, Zhongqiang Huang, Fei Huang, and Kewei Tu. 2022.
\newblock \href {http://arxiv.org/abs/2105.03654} {Improving named entity recognition by external context retrieving and cooperative learning}.

\bibitem[{Yasunaga et~al.(2022)Yasunaga, Leskovec, and Liang}]{yasunaga-etal-2022-linkbert}
Michihiro Yasunaga, Jure Leskovec, and Percy Liang. 2022.
\newblock \href {https://doi.org/10.18653/v1/2022.acl-long.551} {{L}ink{BERT}: Pretraining language models with document links}.
\newblock In \emph{Proceedings of the 60th Annual Meeting of the Association for Computational Linguistics (Volume 1: Long Papers)}, pages 8003--8016, Dublin, Ireland. Association for Computational Linguistics.

\bibitem[{Yuan et~al.(2022{\natexlab{a}})Yuan, Yuan, Gan, Zhang, Xie, and Yu}]{yuan2022biobart}
Hongyi Yuan, Zheng Yuan, Ruyi Gan, Jiaxing Zhang, Yutao Xie, and Sheng Yu. 2022{\natexlab{a}}.
\newblock Biobart: Pretraining and evaluation of a biomedical generative language model.
\newblock \emph{arXiv preprint arXiv:2204.03905}.

\bibitem[{Yuan et~al.(2022{\natexlab{b}})Yuan, Yuan, and Yu}]{yuan2022generative}
Hongyi Yuan, Zheng Yuan, and Sheng Yu. 2022{\natexlab{b}}.
\newblock Generative biomedical entity linking via knowledge base-guided pre-training and synonyms-aware fine-tuning.
\newblock \emph{arXiv preprint arXiv:2204.05164}.

\bibitem[{Yuan et~al.(2021)Yuan, Liu, Tan, Huang, and Huang}]{yuan2021improving}
Zheng Yuan, Yijia Liu, Chuanqi Tan, Songfang Huang, and Fei Huang. 2021.
\newblock Improving biomedical pretrained language models with knowledge.
\newblock \emph{arXiv preprint arXiv:2104.10344}.

\end{thebibliography}
\bibliographystyle{acl_natbib}

\appendix
\section{Example Prediction of Non-canonical Concept Mentions}
\begin{figure*}[!t]
\centering
\includegraphics[width=0.9\linewidth]{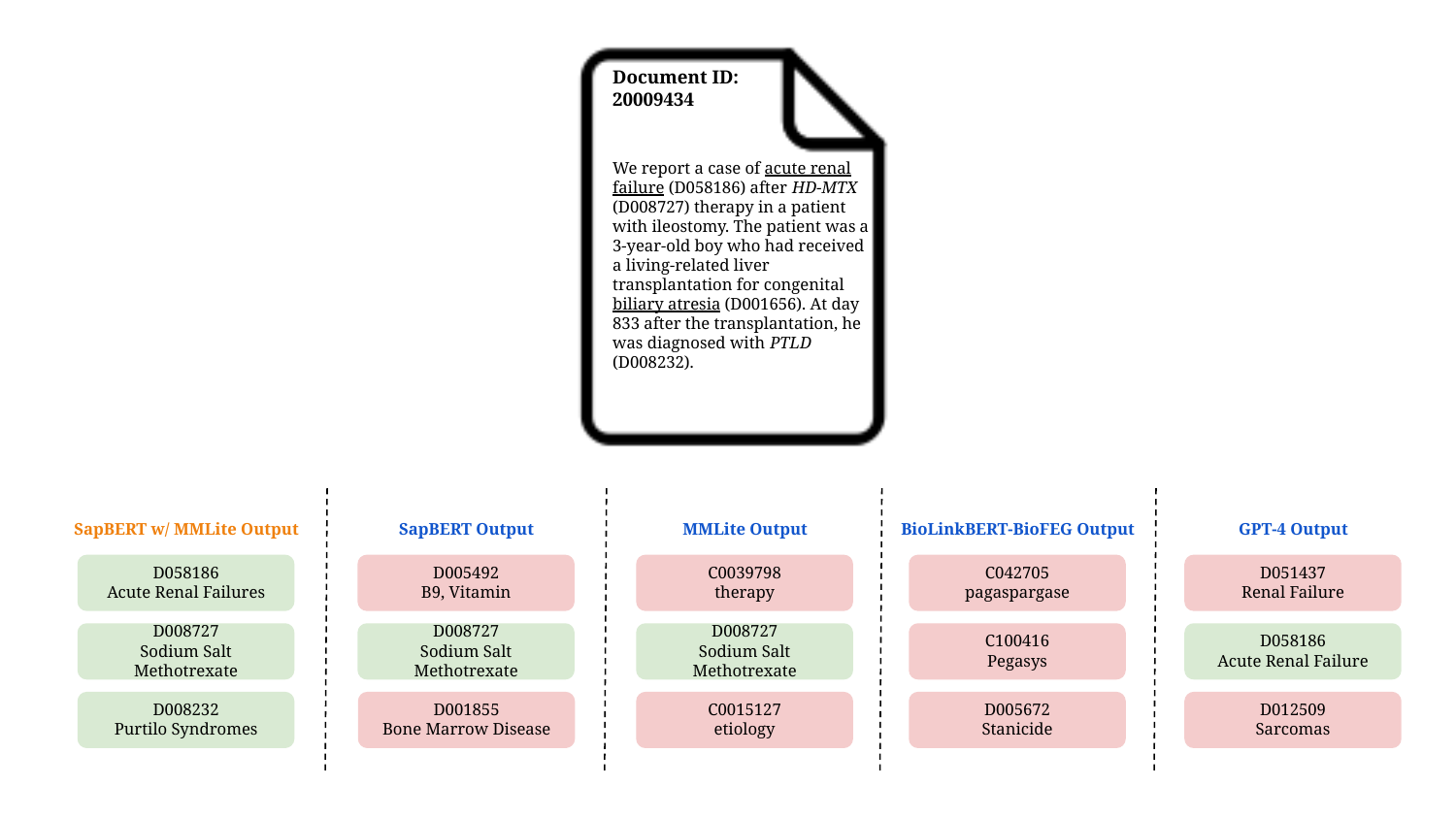}
    \caption{Example of non-canonical concept prediction output. \underline{Underscored} text indicates canonical concept mentions. \textit{Italic} text indicates non-canonical concept mentions. (bracketed) text indicates concept IDs.}
\label{Figure:non_can_pred}
\vspace{-2ex}
\end{figure*}

To compare the output of our data-augmented model against other modes in terms of non-canonical concept mentions, we show in Figure \ref{Figure:non_can_pred} an example from the BC5CDR test set. For brevity, we present the top 3 predicted concepts by each model instead of the top 10, as the remaining predictions within the top 10 are incorrect. Our data-augmented model accurately predicts both non-canonical concept mentions, while other models predict at most one concept correctly. This highlights the effectiveness of learning from a pseudo-annotated corpus and considering the entire context in predicting non-canonical concept mentions.

\section{Additional Details of Datasets} \label{Sec:data_char}
We find that many concepts in the test dataset do not have enough training examples.
Table \ref{Table:data_summary} shows that within the test set of NCBI-Disease, 18.2\% of the annotations are not trained, and 36.3\% of the annotations have fewer than 10 training examples. For BC5CDR, 22.7\% of the test set annotations are untrained, and 39.2\% of the annotations have fewer than 10 training examples. Additionally, many concepts are referred to in the documents using non-canonical names that are not found in the ontologies. 38.4\% of the annotations in NCBI-Disease and 21.0\% in BC5CDR are non-canonical concept mentions. Our data exploration indicates that pre-trained language models perform poorly on these low-resource and non-canonical concepts.

\section{Approximate Concept Search} \label{Sec:faiss}
We use the approximate vector search tool FAISS \cite{johnson2019billion} to select the best matching candidate concepts given a document based on L2-distance. In our approach, we treat the document embedding $f(X)$ as the query vector and the concept embedding $\theta e$ as the candidate vector. FAISS constructs an inverted index containing the approximate concept embeddings as follows:
\begin{equation}
    \theta e \approx q(\theta e) = q_{1}(\theta e) + q_{2}(\theta e - q_{1}(\theta e))
\end{equation}
where $q_1$ is the coarse quantizer and $q_2$ is the fine-grained quantizer. Suppose there are $N$ concept embeddings in the collection. $q_1$ creates a collection of centroids $C$ and groups all the concept embeddings to the nearest centroid, where $|C| \approx \sqrt{N}$.

Once the concept embedding is quantized, FAISS performs a nearest neighbor search by $L_{\text{ADC}}$:
\begin{align}
    L_{\text{ADC}} &= k\text{-argmin}_{i=0:|C| \text{ s.t. } q_{1}(\theta e_{i}) \in L_{\text{IVF}}} L_{q_{1}} \\
    L_{q_{1}} &= \| x - q(\theta e_{i}) \|_{2} \\
    L_{\text{IVF}} &= \rho\text{-argmin}_{c \in C} \| x - c \|_{2}
\end{align}
This process reportedly yields the top $k$ candidate concepts for each document 8.5 times faster than exact vector search.

\section{Concept Text Generation} \label{concept_text}
In our work, we follow an approach similar to \cite{sui2023biofeg}, which embeds concepts using text associated with each concept. The distinction in our method is rather than using one concept name, we use all unique concept names and descriptions that are stored in the KG. We observe that some concept names are merely repetitive variations, differing only in case, or with added hyphens, commas, or spaces, as shown in Table \ref{Table:dup_names}. To address this, we normalize all concept names to lower case, strip out hyphens and commas, and replace consecutive spaces, tabs, or return carriages with a single space. After normalization, duplicates are removed. The remaining concept names and descriptions are then packed into a single text sequence and used as input to the concept encoder, discussed in the next section.

\section{Computation Resources}
We fine-tune the transformer models on both manually labeled data and augmented data using a single A100 GPU with 40GB of memory. The fine-tuning process takes 3 hours for the NCBI-Disease dataset and 9 hours for the BC5CDR dataset.

\begin{table}
\centering
\scalebox{0.7}{
\begin{tabular}{lcc}
\toprule
      &  NCBI-Disease & BC5CDR \\
      \midrule
       Concept Source & MeSH+OMIM & MeSH \\
      Training Documents & 592 & 500 \\
      Validation Documents & 100 & 500 \\
      Test Documents & 100 & 500 \\
        Unique concepts & 790 & 2350 \\
      Concept Mentions & 6892 & 28811 \\
      Semantic Types & 10 & 36 \\
      \% untrained concept anno. & 18.2 & 22.7 \\
      \% undertrained concept anno. & 36.3 & 39.2  \\
      \% non-canonical concept mentions & 38.4 & 21.0 \\
        Max unique concepts per doc. & 12 & 22 \\
        Min unique concepts per doc. & 1 & 1 \\
        Avg. unique concepts per doc. & 2.3 & 4.6 \\
        \bottomrule
\end{tabular}}
\caption{\label{Table:data_summary}
Dataset Statistics.}
\end{table}

% \begin{table}
% \centering
% \begin{tabular}{lc}
% \toprule
%      Filter &  \% of Filtered Examples\\
%       \midrule
%       False Abv. & 5.47\\
%       Overlap & 8.5\\
%       Diversity & 34.56\\
%         \bottomrule
% \end{tabular}
% \caption{\label{Table:filtered}
% Percentage of PubMed documents removed by the filters}
% \end{table}

\begin{table}
    \centering
    \resizebox{\columnwidth}{!}{
    \begin{tabular}{lcc}
        \toprule
        Variation Type & Original Concept Name & Varied Duplicate Name\\
        \midrule
        Inversion & Polyposis Coli & Coli, Polyposis\\
        Hyphen & T-Cell & T Cell \\
        Case & Chromosomal Disorder & chromosomal   disorder\\
        \bottomrule
    \end{tabular}}
    \caption{\label{Table:dup_names}
    Various forms of duplicate concept names.}
\end{table}

\section{Hyperparameters}
We use the Huggingface pre-trained models on Huggingface for BioBERT~\footnote{\url{https://huggingface.co/dmis-lab/biobert-base-cased-v1.2}}, BiomedBERT~\footnote{\url{https://huggingface.co/microsoft/BiomedNLP-BiomedBERT-base-uncased-abstract-fulltext}}, BioLinkBERT~\footnote{\url{https://huggingface.co/michiyasunaga/BioLinkBERT-base}} and SapBERT~\footnote{\url{https://huggingface.co/cambridgeltl/SapBERT-from-PubMedBERT-fulltext}}. The initial learning rate is set at 0.0001 and decays throughout the training epochs. We maintain a batch size of 16 and fix the InfoNCE temperature at 1. For the rest of the hyperparameters, we use the default values provided in the Huggingface checkpoints. 

% additional case study of non canonical concept predictions of different models

%\label{sec:appendix}

%This is a section in the appendix.

\end{document}